# Hierarchical Framework for Interpretable and Probabilistic Model-Based Safe Reinforcement Learning⋆

Ammar N. Abbas[a,b,∗], Georgios C. Chasparis[a] and John D. Kelleher[c]

[a]*Data Science, Software Competence Center Hagenberg, Softwarepark 32a, Hagenberg, 4232, Austria*
[b]*Department of Computer Science, Technological University Dublin, Dublin, D02 HW71, Ireland*
[c]*ADAPT Research Centre, Technological University Dublin, Dublin, D02 HW71, Ireland*

## ABSTRACT

The difficulty of identifying the physical model of complex systems has led to exploring methods that do not rely on such complex modeling of the systems. Deep reinforcement learning has been the pioneer for solving this problem without the need for relying on the physical model of complex systems by just interacting with it. However, it uses a black-box learning approach that makes it difficult to be applied within real-world and safety-critical systems without providing explanations of the actions derived by the model. Furthermore, an open research question in deep reinforcement learning is how to focus the policy learning of critical decisions within a sparse domain. This paper proposes a novel approach for the use of deep reinforcement learning in safety-critical systems. It combines the advantages of probabilistic modeling and reinforcement learning with the added benefits of interpretability and works in collaboration and synchronization with conventional decision-making strategies. The BC-SRLA is activated in specific situations which are identified autonomously through the fused information of probabilistic model and reinforcement learning, such as abnormal conditions or when the system is near-to-failure. Further, it is initialized with a baseline policy using policy cloning to allow minimum interactions with the environment to address the challenges associated with using RL in safety-critical industries. The effectiveness of the BC-SRLA is demonstrated through a case study in maintenance applied to turbofan engines, where it shows superior performance to the prior art and other baselines.

## 1. Introduction

Data-driven decision-making in the current industrial revolution of industry 4.0 is considered an open research topic as more advanced methodologies keep on emerging to make it more optimal even in very complex environments. There are commonly three approaches that are used while making decisions in a safety-critical environment, (i) model-based (relying on physical model), (ii) knowledge-based (relying on expert knowledge), and (iii) data-driven (relying on data analytics and machine learning algorithms), as mentioned by Bousdekis et al. (2019).

There are several challenges associated with the use of RL in safety-critical systems as also mentioned by Wuest, Weimer, Irgens and Thoben (2016); Ibarz, Tan, Finn, Kalakrishnan, Pastor and Levine (2021).:

1. Training of DRL requires continuous interaction with the environment to be able to understand the system dynamics and its reward model. For safety-critical industries, it is either expensive or impossible to have such random interactions with the environment.
2. Industries tend to rely more on white box systems that are more interpretable and human-understandable rather than black box methodologies that do not guarantee convergence or have higher uncertainty as indicated by Khan and Khan (2012).

⋆This publication is the result of the research and activities done along the Collaborative Intelligence for Safety-Critical systems (CISC) project; which has received funding from the European Union's Horizon 2020 Research and Innovation Program under the Marie Skłodowska-Curie grant agreement no. 955901. The research reported in this paper has been performed within the frame of SCCH, part of the COMET Program managed by FFG. The work of Kelleher is also partly funded by the ADAPT Centre which is funded under the Science Foundation Ireland (SFI) Research Centres Program (Grant No. 13/RC/2106_P2).

The proposed hybrid framework addresses the major challenges encountered while applying RL to real-world and safety-critical systems. It combines the advantages of probabilistic modeling with reinforcement learning.

∗Corresponding author
✉ ammar.abbas@scch.at (A.N. Abbas); georgios.chasparis@scch.at (G.C. Chasparis); john.d.kelleher@tudublin.ie (J.D. Kelleher)
ORCID(s): 0000-0002-2578-5137 (A.N. Abbas); 0000-0003-3059-3575 (G.C. Chasparis); 0000-0001-6462-3248 (J.D. Kelleher)

3. Entirely replacing conventional methodologies with Machine Learning (ML) or RL techniques in such critical scenarios is not recommended due to the uncertain nature of such methods.
4. DRL is known to be sample and data inefficient and training becomes intractable once it enters complex and high-dimensional state-action spaces.

RL has been shown to be an effective method for solving decision problems that involve time series data, such as predictive maintenance Skordilis and Moghaddass (2020a). The advantage of RL is that it allows for learning in real-time, without the need for pre-existing data sets Sutton and Barto (2018). However, when the decisions to be made by the RL agent are relatively rare in the data set, the resulting policy may be influenced by irrelevant factors, leading to suboptimal performance. Additionally, the optimal policy derived by RL does not provide insights into the underlying causes of the decision, which limits the ability of humans to collaborate with the RL system. Furthermore, in real-world industrial settings, RL is typically applied to raw sensor data, which may not provide enough information about the underlying factors that influence the decision-making of the system, such as its health, which can further limit the performance of the RL agent.

In this paper, we try to address these challenges and propose a novel general methodology of DRL for safety-critical systems based on the hierarchical framework for interpretable and probabilistic model-based safe reinforcement learning. We name the proposed architecture a *Behavioral Cloning-Based Specialized Reinforcement Learning Agent (BC-SRLA)*. It combines the advantages of a probabilistic model with reinforcement learning in a hierarchical framework, which incorporates a training method that initializes the baseline policy to have minimum interactions with the environment. We have further enhanced it with an interpretable policy in a human-understandable form enabling a collaborative environment. BC-SRLA does not replace the conventional methodology, but reinforces it with the added benefit of DRL methods and is activated once required. Moreover, the data inefficiency is targeted with the specialization of the state where the need of DRL is mostly required, such as in abnormal situations or when the system is near-to-failure.

The proposed architecture is inspired by *Hierarchical Reinforcement Learning (HRL)* and its benefits in long-horizon sparse reward environments as discussed by Pateria, Subagdja, Tan and Quek (2021). The use of hierarchical modeling is a solution to the problem of sample-inefficiency in RL, which can be addressed by decomposing the large state space of long-horizon tasks into several specialized short-horizon tasks. However, as opposed to having hierarchies of RL agents, a hierarchical structure is introduced here as a result of the introduction of a supervisory probabilistic model for monitoring the operating conditions of the process. The proposed method involves two steps: first, a probabilistic model is used to filter large amounts of non-relevant data generated during normal operation and detect states in which critical abnormality is imminent. In the second step, a DRL agent learns the optimal policy based on these critical states. The probabilistic (supervisory) model has access to the full information. It works at the higher abstraction level that segments and infers with respect to the operating state of the system. On the other hand, at the lower level, RL accesses only the particular state of the system where the probabilistic model identifies its necessity and is specialized to only that given state. The idea is inspired by human behavior, where the mind identifies the situation in which a particular action needs to be taken or how to act in an abnormal situation and then the lower-level functions of muscle movements are activated accordingly. This helps RL to be sample efficient and increases its exploration efficiency as the state-action space is reduced to a particular scenario.

Our experimental results indicate that this state-/event-based approach with dynamic data pre-filtering has comparable performance to prior methods that train RL agents on the full data set, while also increasing training efficiency and allowing for more interpretable policies. The probabilistic model is used to learn the state representation of the system and the DRL is used to model the state-action pairs of the environment. We evaluate our approach using the NASA C-MAPSS (Commercial Modular Aero-Propulsion System Simulation) turbofan degradation data sets which consist of multivariate time series sensor readings and operating conditions based on flight cycles within a run-to-failure simulation.

***Structure:*** The paper is divided into the following format: Section 2 addresses the related work in the field of applying RL to safety-critical decision-making applications and briefly discusses the literature on HRL. This section also includes the contribution of the paper. Section 3 illustrates the proposed architecture and discusses its components. Section 3.3, further, reframes the proposed methodology to a specific model and redefines its elements. The specific type of probabilistic model and the utilized RL architecture are further defined in that section along with the interpretability factor. Section 4 frames predictive maintenance as an RL problem and further models the environment dynamics, reward formulation, and evaluation criteria for the given case study. Section 5 discusses the setup for experimentation and the baselines for comparison. Moreover, the experiment related to the model's hyperparameter search is also identified in that section. Section 6 compares and evaluates the proposed architecture with the baseline and the prior art and Section 7 defines the second part of the experiments related to the interpretations.

## 2. Related Work and Contribution

**Reinforcement Learning in Safety-Critical Industrial Applications:**

Manufacturing involves tasks that require decision-making by plant operators, including scheduling, process control, and monitoring. These tasks can be complex and require expert knowledge and programming time to be performed efficiently. Automating these tasks is a complex optimization problem that requires the use of novel technologies, such as machine learning (ML). There are several requirements for an ML application in the manufacturing industry, including the ability to handle high-dimensional problems and datasets with moderate effort, limited computational capacity, simplify potentially complex outputs, adapt to changes in the environment in a cost-effective manner, expand on previous knowledge through experience, work with available manufacturing data without needing very detailed information, and discover relevant relationships within and between processes that are discussed by Chien, Dauzère-Pérès, Huh, Jang and Morrison (2020); Wuest et al. (2016); Morgan, Halton, Qiao and Breslin (2021).

Among the many machine learning paradigms, reinforcement learning is suitable for tasks such as those in industrial processes. The trial-and-error learning through the interaction with the environment and not requiring pre-collected data and prior expert knowledge allows RL algorithms to adapt to uncertain conditions, which is also discussed by Panzer and Bender (2022). Some applications can be found in manufacturing, for instance, in scheduling tasks as an example demonstrated by Dong, Xue, Xiao and Li (2020), maintenance as a case study researched by Rodríguez, Kubler, de Giorgio, Cordy, Robert and Le Traon (2022); Yousefi, Tsianikas and Coit (2022), process control described by the authors Spielberg, Tulsyan, Lawrence, Loewen and Gopaluni (2020), energy management example elaborated by Lu, Li, Li, Jiang and Ding (2020), assembly task mentioned by Tortorelli, Imran, Delli Priscoli and Liberati (2022), and robot manipulation that in detail has been discussed by Beltran-Hernandez, Petit, Ramirez-Alpizar and Harada (2020); Schoettler, Nair, Luo, Bahl, Ojea, Solowjow and Levine (2020).

The combination of Deep Neural Networks (DNN) and Reinforcement Learning (RL) has emerged into an exciting field of Machine Learning known as *Deep Reinforcement Learning (DRL)*. Recent ground-breaking results from the state-of-the-art RL algorithms have proved to rule out humans in multiple complex games, such as chess, Go, Atari, etc as demonstrated by Silver, Hubert, Schrittwieser, Antonoglou, Lai, Guez, Lanctot, Sifre, Kumaran, Graepel et al. (2018); Silver, Huang, Maddison, Guez, Sifre, Van Den Driessche, Schrittwieser, Antonoglou, Panneershelvam, Lanctot et al. (2016); Mnih, Kavukcuoglu, Silver, Rusu, Veness, Bellemare, Graves, Riedmiller, Fidjeland, Ostrovski et al. (2015); Mnih, Kavukcuoglu, Silver, Graves, Antonoglou, Wierstra and Riedmiller (2013). However, as fast as DRL is enhancing its capabilities to master its application in games, the gap between its real-world, safety-critical systems is becoming wider. There are comparatively few studies, where DRL has been implemented on safety-critical industrial cases, some of which are presented by Rodríguez et al. (2022); Senthil and Sudhakara Pandian (2022); Spielberg, Gopaluni and Loewen (2017); Shin, Badgwell, Liu and Lee (2019).

The use and innovation of DRL have evolved in Industry 4.0 and 5.0, specifically in the manufacturing industry. However, it is still challenging to implement DRL in real-world cases from simulation environments. A proposed solution cited in some of the literature is the use of digital twins. Authors from del Real Torres, Andreiana, Ojeda Roldán, Hernández Bustos and Acevedo Galicia (2022); Panzer and Bender (2021a) have provided a systematic literature review on the applications of DRL in various sectors of industrial production systems, such as assembly, process control, robotics, scheduling, maintenance, quality control, and energy management. The authors have further highlighted that DRL algorithms are data-driven and can be reconfigured to meet industry process needs, and are being implemented across the activities of the manufacturing industry, outperforming traditional techniques and increasing overall resilience and adaptability. However, the lack of a general framework is highlighted for each application, and the drawback of having a domain-focused methodology that is not scalable. These are one of the major challenges that need to be tackled to enable the widespread adoption of deep RL in production systems.

Several hybrid architectures are presented in the literature that provides the usage of multiple Machine Learning methods to work collaboratively to have the added benefits of each of the models. Such approaches also include the use of RL in an interactive environment where the model learns from both its own experience by interacting with the environment and observing rewards as well as from the experience of a human expert through a human-in-the-loop mechanism. The authors Lepenioti et al. (2020) discussed the use of predictive and prescriptive analytics using Machine Learning (ML) approaches in smart manufacturing, using a predictive maintenance strategy in the steel industry as a case study. Two different ML algorithms, Recurrent Neural Networks (RNN) and Multi-Objective Reinforcement Learning (MORL) were selected for each analytics and used collaboratively. Their solution tries to target the difficulty of creating a generalizable architecture for all industrial processes while providing interpretation and explainability of the analysis.

**Hierarchical Reinforcement Learning:**

Although the proposed architecture does not relate to the hierarchical structure of reinforcement learning agents, it presents a hierarchical framework that uses two different machine learning approaches, namely, probabilistic modeling and reinforcement learning. The reason to include a review of the literature for Hierarchical Reinforcement Learning (HRL) is that it resembles the methodology discussed in this paper to some extent in terms of its features, advantages, scalability, and concepts. HRL consists of two broader branches, (i) Feudal and (ii) Option. Feudal presents a "manager and sub-manager" hierarchy, whereas option relates to implementing a specific RL model to the given situation. Our proposed architecture incorporates the perks of both the architecture in terms of having a higher abstraction model as feudal RL using the probabilistic approach and in other ways having option RL for choosing the correct RL model for the autonomously identified states as well as providing interpretations for the given action or for the given identification of the state. A survey by Pateria et al. (2021) provides an overview of the various HRL approaches and their applications, based on a novel taxonomy. Five broad classes of HRL methods have been identified that address the issues of learning hierarchical policies, subtask discovery, transfer learning, and multi-agent learning. Additionally, a set of open problems is outlined to help further the research of HRL. These open problems include lifelong skill discovery and utilization, increasing data efficiency by leveraging high-level planning, and providing theoretical guarantees of optimality.

A similar hierarchical structure is proposed by Skordilis and Moghaddass (2020b) utilizing deep reinforcement learning and particle filtering as a framework. It generates real-time control and maintenance policies and estimate remaining useful life for sensor-monitored degrading systems. They demonstrated the effectiveness of these methods through numerical experiments using simulated data and a turbofan engine dataset. The authors focus on using raw sensor data directly to make the algorithm generalizable for real-time scenarios. They proposed a Bayesian Network-based Deep Reinforcement Learning approach where degradation states are defined as latent states, which are not directly observable. The Bayesian framework is used to map the raw sensor data into belief states that serve as input to the DRL network to derive the optimal action policies.

## 2.1. Contribution

As discussed above, multiple RL approaches are indicated for various industrial applications, however, to the best of our knowledge, most of the research focuses on the case study in hand and fails to provide a general methodological architecture that can be adopted across different industrial tasks or safety-critical systems. Therefore, the primary contribution of this research is to identify a general framework and to address the major challenges encountered while applying RL to real-world, safety-critical systems. The framework has these specific features:

- Segmentation of the state-space through probabilistic modeling.

- Initializing baseline policy for guided exploration using Behavioral Cloning (BC).

- Identifying a particular state or an abnormal state with the help of segmented state-space through probabilistic modeling and state value function from RL.

- Generation of interpretations of the operating condition through a combination of Human-in-Loop (HIL) feedback and by fusing information from the probabilistic model and RL state-action value function.

In this paper, we have proposed a novel approach called Behavioral Cloning-Based Specialized Reinforcement Learning Agent (BC-SRLA) for using deep reinforcement learning in safety-critical systems. The BC-SRLA combines probabilistic modeling and reinforcement learning with interpretability and works in collaboration with conventional decision-making strategies. It is activated in specific situations identified autonomously through the probabilistic model and reinforcement learning and is initialized with a baseline policy using policy cloning to minimize interactions with the environment. The effectiveness of the BC-SRLA is validated through a case study in maintenance applied to turbofan engines. The paper addresses the challenges associated with using RL in safety-critical systems, such as the need for continuous interaction with the environment, reliance on interpretable and human-understandable systems, and uncertainty in the RL methods.

## 3. Methodology
### 3.1. Preliminaries
#### 3.1.1. Probabilistic Modeling

Probabilistic modeling is a way of representing and reasoning about uncertain events or observations using mathematics, as denoted by Brémaud (2012). It involves building a model that estimates the probability of different outcomes or events based on certain assumptions or observed data. This type of modeling has many applications, including predicting the likelihood of outcomes in scientific or industrial contexts, understanding the factors that influence the likelihood of events, and making decisions in uncertain situations.

There are several types of probabilistic models, including Bayesian networks, Markov models, hidden Markov models, and Gaussian mixture models. Bayesian networks are graphical models that show the probabilistic relationships between variables or events and can be used to make predictions or decisions with incomplete data. Markov models describe a sequence of events or states, where the probability of transitioning from one state to another depends only on the current state. They are often used to model processes that change over time, such as the spread of diseases or customer behavior. Hidden Markov models are a type of Markov model that is used to model systems where the state is not directly observable, but can be inferred from observations. These models are commonly used in speech recognition and natural language processing. Gaussian mixture models are probabilistic models used for clustering, where a dataset is thought to be generated by a mixture of different underlying distributions. They are frequently used in machine learning and data analysis.

Probabilistic modeling involves several steps, including defining the variables and events of interest and specifying their probabilistic relationships, selecting a suitable model family and defining its parameters, fitting the model to the data using methods such as maximum likelihood estimation or Markov chain Monte Carlo sampling, and using the fitted model to make predictions or decisions based on uncertain data. Probabilistic modeling can be a powerful tool for understanding and predicting uncertain events and has numerous applications in fields of industrial applications, such as anomaly detection, decision-making, abnormality in process control, maintenance, etc.

#### 3.1.2. Reinforcement Learning

Reinforcement Learning (RL) has shown promising results when applied to stochastic decision-making Skordilis and Moghaddass (2020a). RL systems learn by interacting with an environment and at each state observe the rewards received for each action with the optimization objective to maximize the total cumulative reward. The problem formulation is based on the concept of the Markov Decision Process (MDP); solved via Dynamic Programming (DP), which is the mathematical modeling of decision-making under stochasticity, as mentioned by Sutton and Barto (2018). However, in a complex environment where the state and action spaces are either continuous or very large, it is impossible to store the values for each state-action pair in the memory. Therefore, recent development in Deep Learning (DL) has led to the development of Deep Reinforcement Learning (DRL) that enables to use of the concept of function approximation, which generalizes effectively to enormous state-action spaces through the approximation of unobserved states, which is also stated by Bertsekas and Tsitsiklis (1996).

*Q-Learning* Q-learning is a reinforcement learning algorithm, first proposed by Watkins and Dayan (1992). It is used to learn the optimal action-selection policy for a given environment. The goal of Q-learning is to learn a function called the Q-function, which gives the expected return (or reward) for each action at each state. The equation for one of the algorithms of reinforcement learning known as Q-learning can be represented as shown in Equation (1):

$$Q(s,a) \leftarrow Q(s,a) + \alpha \left[ r + \gamma \max_{a'} Q(s',a') - Q(s,a) \right] \quad (1)$$

Here, $Q(s,a)$ is the current estimate of the optimal state-action value function for a given state $s$ and action $a$, $\alpha$ is the learning rate, $r$ is the reward received after taking action $a$ in state $s$, $\gamma$ is the discount factor, and $s'$ and $a'$ are the next state and action, respectively. The term $\max_{a'} Q(s',a')$ represents the maximum action value of all possible actions in the next state $s'$. The equation updates the current estimate of the state-action value function by moving it closer to the target value, which is the sum of the reward and the discounted maximum action value of the next state. This update helps the agent learn the optimal state-action value function, which can be used to select the best action in each state and maximize the cumulative reward over time.

DRL in the context of decision-making is not able to provide interpretations or the root cause of the anomaly, and therefore, it keeps the human out of the loop in such decision-making tasks. Due to this, the use of collaborative intelligence is limited and human experts can not supervise the model to learn. Furthermore, in industrial environments, it focuses on learning directly from the observed raw sensor data that does not provide information about the unobserved hidden parameters of the system such as its health, which limits the agent to behave sub-optimally. Probabilistic modeling as discussed earlier can overcome the challenges faced by DRL through (i) introducing a temporal structure in the model, that is lost with DRL due to maintaining independent and identically distributed characteristics of DL Skordilis and Moghaddass (2020a), (ii) learning unobserved hidden states and interpretation, and (iii) reducing the input variables and complexity of the raw data Yoon, Lee and Hovakimyan (2019). Therefore, the probabilistic model learns the hidden state representation of the system and the DRL constructs the state-action pair modeling of the environment.

### 3.1.3. Behavioral Cloning

Behavioral Cloning (BC) Michie, Bain and Hayes-Miches (1990) represents a term that defines an approach to imitate the behavioral response of an expert for a given situation and try to mimic it. It is used in industries in the context of robotics as well as in other disciplines to try and mimic human responses or responses of a conventional controller. BC helps RL to learn an initial/base policy around which the RL agent can explore and optimize. In safety-critical industries, this can help RL agent to avoid random exploration for minimizing costs and catastrophic failures.

The loss function for BC as also defined by Goecks, Gremillion, Lawhern, Valasek and Waytowich (2020), through the mean squared error can be represented as:

$$\mathcal{L}_{BC}(\theta_\pi) = \frac{1}{2}\left(a_t^A(s_t \mid \theta_\pi) - a_t^E\right)^2 \qquad (2)$$

where $\theta_\pi$ represents the network parameters for a behavior policy, $a_t^A(s_t \mid \theta_\pi)$ is the behavior policy, $s_t$ is the current state, and $a_t^E$ is the expert behavior.

## 3.2. Proposed Architecture

Behavioral Cloning-Based Specialized Reinforcement Learning Agent (BC-SRLA) is a hierarchical framework composed of combining probabilistic modeling and Reinforcement Learning (RL). The methodology is illustrated in Figure 1. The green arrows in the figure represent the processes involved in the pretraining and training phase of the architecture. The red arrows define the processes that are involved during the inference phase of the overall methodology. Finally, the black arrow ($\pi^*$) represents the output of the architecture as the optimal policy which is the combination of both the RL policy and the expert action synchronized together. This synchronization is in effect during the training phase as well as during the inference. The probabilistic model acts as the higher hierarchy that defines the categorization of the different states of the system in which it can and clusters them for RL specialization and interpretations (such as through feature importance). The lower part of the architecture hierarchy consists of an RL agent that is pretrained through policy cloning on the expert's strategy to initialize a baseline policy. Moreover, the RL agent is only activated for training and inference, if required, for example in case of anomaly detection. This helps the RL agent to be trained on a specific problem set with reduced state-action exploration space. Detection of these particular required states is performed by the upper hierarchy as mentioned earlier, which passes that information to the lower level. RL agent then provides suggestions to the operator based on the decisions or actions to act upon. RL agent also considers the expert's strategy and tries to synchronize its action accordingly. Furthermore, the state value function $V$ of the pretrained RL agent helps define the labels for the state clustered by the probabilistic model, such as an abnormal state would result in lower values of $V$. The state-action value function $Q$ of the trained RL agent helps with the interpretation to humans in combination with the probability output to explain the suggested/taken actions or decisions. The expert also plays a vital role in the interpretations to help train the model for defining the interpretations in a more formulated approach that takes into account further evidence through multiple aspects of the system. Moreover, the expert trajectory acts as a safety net for the methodology, which is necessary for safety-critical environments to avoid abnormal behavior of the RL agent for which it is not trained on.

### 3.2.1. Assumptions

These particular assumptions are taken into account:

- The Probabilistic model is trained perfectly and provides the exact state information to the RL agent.

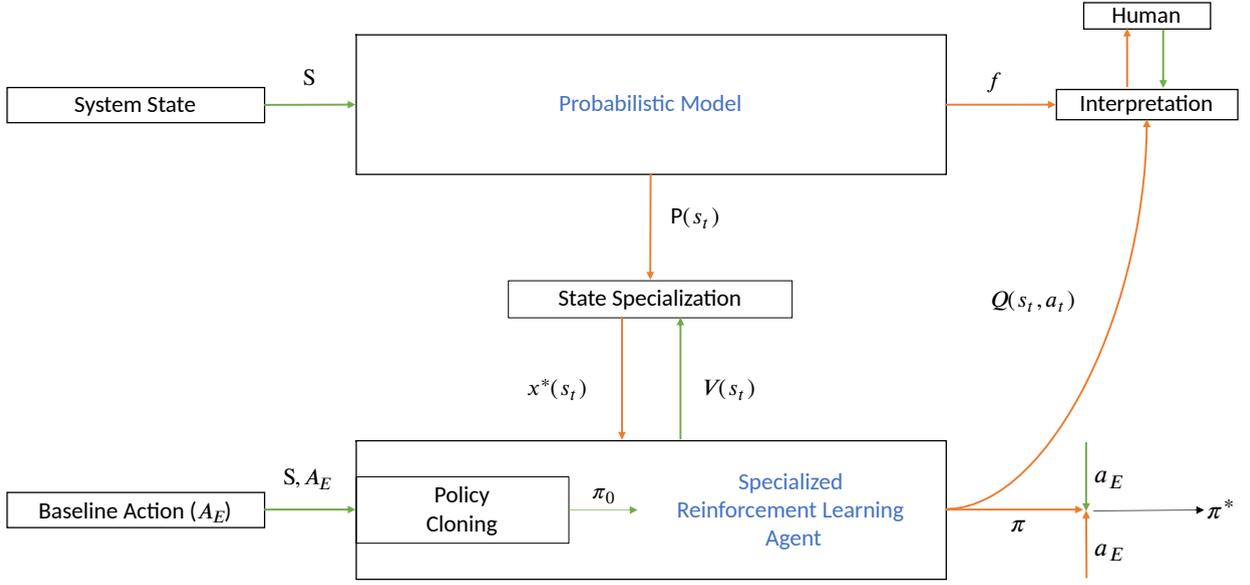

**Figure 1:** Behavioral Cloning-Based Specialized Reinforcement Learning Agent (BC-SRLA)

- The data from the simulator/environment is synchronized without any missing data.
- There is no lag between the system state information input and the transmission of the action to the environment.

### 3.3. Framed Methodology:
### Input-Output Hidden Markov Model-Based Deep Reinforcement Learning (IOHMM-DRL)

The specific methodology derived from the general architecture from Section 3.2, which will be used in this paper is a hierarchical model integrating an Input-Output Hidden Markov Model (IOHMM) and Deep Reinforcement Learning (DRL). Within this hierarchical model, the IOHMM (used as the probabilistic model) aims to identify when the system is approaching a desired (such as failure) state. Once the IOHMM has entered this specific state, the task of the DRL agent (used as the reinforcement learning model) is to optimize the decision of when to take the desired action (Such as replacing the equipment to maximize its total useful life). This IOHMM-DRL model allows for the state- or event-based optimization. This further allows for more efficient DRL training, since the training data set is restricted to the imminent-to-failure states. Such agents can be deployed under situation-dependent adaptations as mentioned in Panzer and Bender (2021b). Figure 2 illustrates the proposed hierarchical model which we name Input-Output Hidden Markov Model-Based Deep Reinforcement Learning (IOHMM-DRL).

The DRL training and optimization process is relatively standard. We use Deep Learning (DL) as a function approximator that generalizes effectively to enormous state-action spaces through the approximation of unvisited states Bertsekas and Tsitsiklis (1996) as shown in Equation (3). In this equation $L_i$ denotes the loss function, $y_i$ is the TD target; which is the sum of the observed one-step reward and the discounted next Q (action) value conditioned on the current state and action, $Q(s,a)$ is the estimation of the Q value of the current state-action pair parameterized by $\theta$.

$$L_i(\theta_i) = \mathbb{E}_{a \sim \mu}\left[(y_i - Q(s,a;\theta_i))^2\right];$$
$$y_i := \mathbb{E}_{a' \sim \pi}\left[r + \gamma \max_{a'} Q(s',a';\theta_{i-1}) \mid S_t = s, A_t = a\right] \quad (3)$$

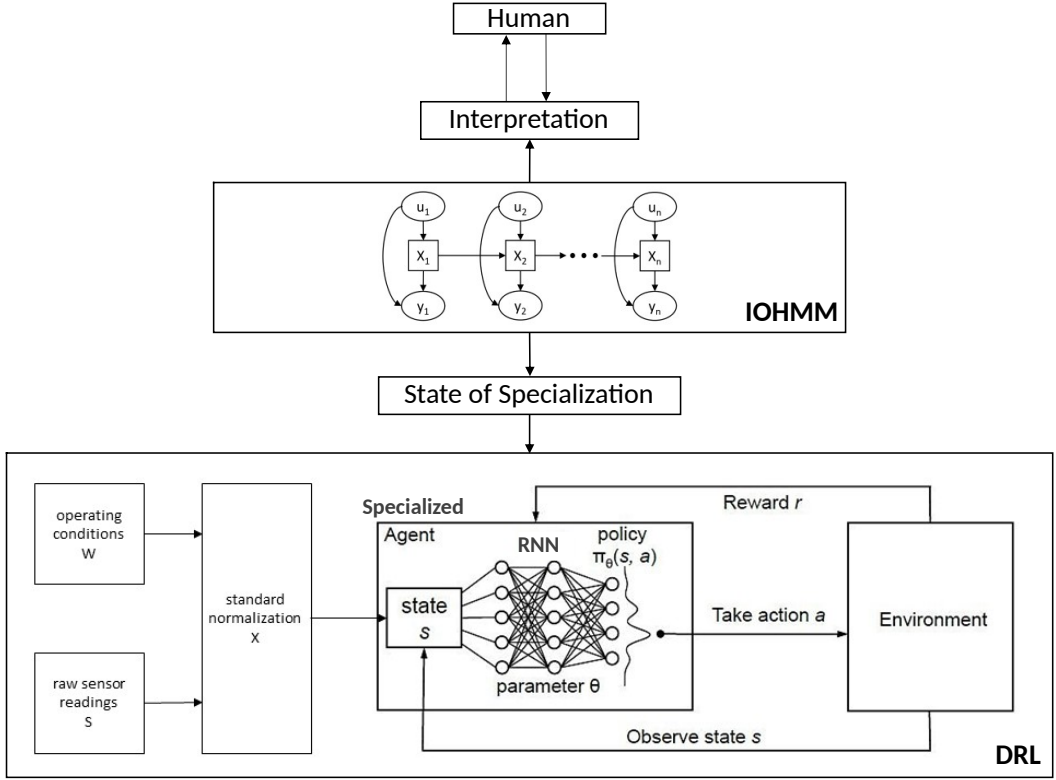

Figure 2: Input-Output Hidden Markov Model-Based Deep Reinforcement Learning (IOHMM-DRL) framework.

At a high level, an IOHMM is used that is an extension to a standard HMM model Bengio and Frasconi (1995). In a standard HMM model (as described Rabiner (1989)) the training optimization objective is to identify the model parameters that best determine the given sequence of observations. To predict the probability of being in a particular hidden state, given the observation sequence $Y$ and trained model parameters $\lambda$ (initial state, transition, and emission probability matrices), Equation 4 is used. $\gamma$ is the vector defining the probability of being in each hidden state at a particular time, which will be used as the input to DRL in our baseline extension. Equation 5 predicts the most probable hidden state that in this context leads to the health degradation state, given the sequence of sensor observations. However, this does not provide information on the most probable sequence of states; as it might be possible that the most probable state at a particular time step may not be the most optimal state sequence, given the history. This problem is solved by the Viterbi algorithm Forney (1973) as shown in Equation 6, where, in this context, $\delta$ is used to predict the health degradation state sequence of the equipment, where the last cycle of each equipment determines the failure state.

$$\gamma_t(i) = P\left(x_t = S_i \mid Y, \lambda\right) \tag{4}$$

$$x_t = \underset{1 \leq i \leq N}{\operatorname{argmax}} \left[\gamma_t(i)\right], \quad 1 \leq t \leq T \tag{5}$$

$$\delta_t(i) = \max_{x_1,\cdots,x_{t-1}} P\left[x_1 \cdots x_t = i, Y_1 \cdots Y_t \mid \lambda\right] \tag{6}$$

One of the limitations of HMM is that the mathematical model does not take into account any input conditions that affect the state transition and the emission probability distribution of the observations (outputs). In the context of industrial settings, these inputs are the operating conditions that heavily influence the system's state and control the system's behavior. Therefore, IOHMM is used to have a more general model architecture that can utilize the information of operating conditions, which modifies Equation (4 and 6) to Equation (7 and 8) with $\lambda$ being conditioned on the input ($U$) as well.

$$\gamma_t(i) = P\left(x_t = S_i \mid U, Y, \lambda\right) \tag{7}$$

$$\delta_t(i) = \max_{x_1,\cdots,x_{t-1}} P\left[x_1 \cdots x_t = i, Y_1 \cdots Y_t \mid U, \lambda\right] \tag{8}$$

### 3.4. Pseudocode

The inference algorithm for the IOHMM-DRL is described in Algorithm 1.

---

**Algorithm 1** Input-Output Hidden Markov Model-Based Deep Reinforcement Learning (IOHMM-DRL)

---

*STEP I:* IOHMM Training
**Input:**
$n$: number of hidden states
$Y$: output sequences
$U$: input seauences
**Output:** $\lambda$: model parameters (initial, transition, and emission probability)

*STEP II:* Viterbi Algorithm (IOHMM inference)
**Input:** $\lambda$, $U$, $Y$
**Output:** $\delta_t(i) = \max_{x_1,\cdots,x_{t-1}} P\left[x_1 \cdots x_t = i, u_1 \cdots u_t, y_1 \cdots y_t \mid \lambda\right]$

*STEP III:* DRL Training
**Input:**
$\delta_s$: specific event (such as failure)
$S_t$: $u_t + y_t$
Environment modeling
Deep reinforcement learning: Algorithm A.2 of Appendix A
**Output:** $\hat{Q}^*(S_t, A_t)$

*STEP IV:* IOHMM-DRL Inference
**Input:** $\lambda$, $\hat{Q}^*(S_t, A_t)$, $S_t$: $(U_t, Y_t)$
Step II, Algorithm 2 for interpretations
$\delta \rightarrow$ Specialized state $(X_s) \rightarrow U_s, Y_s$
**if** $S_t$ in $X_s$ **then**
    $\hat{Q}^*(s_t, a_t)$
    Perform action in the environment
    Observe next state and reward
**else**
    $a_t$ = default action
**end if**
**Output:** $\hat{Q}^*(\delta_t, s_t, a_t)$

---

### 3.5. Interpretability with IOHMM

Beyond the performance considerations of the model, the IOHMM component can provide a level of interpretability in terms of identifying critical operating conditions/states, such as failure states, the root cause of failure, and stages of health degradation. Based on the state sequence distributions predicted by the IOHMM from Equation (8), each state of a particular event can be decoded, such as the failure mode or degradation stage, as shown in Giantomassi et al. (2011). To discover the most relevant sensor readings corresponding to these abnormal conditions, such as failure states that triggered the IOHMM to predict such a state, feature importance is performed that leads to the root cause analysis of the anomalous behavior. In such an approach, raw sensor readings are used as the input feature for the model and IOHMM state predictions are used as the target. After fitting the model, the importance of each sensor can be extracted for each IOHMM state. Algorithm 2 defines the feature importance usage in the context of IOHMM state interpretation.

---
**Algorithm 2** Feature Importance
> **Input:**
> Viterbi state predictions as target classes: $\hat{S}$
> Normalized sensor readings as features: $X$
> **repeat**
>     Fit features with the classes
>     Extract the relevance of features corresponding to every class
>     **Output:** Feature relevance
> **until** all features are evaluated for concerned states
---

In the context of maintenance applications, apart from the failure event hypothesis, it is necessary to measure the health state of the equipment at different points to generate an alarm for the user when the equipment reaches a critical point in its lifetime. The interpretations are based on the critical points along the equipment degradation curve as shown in Figure 3 and the range of observed IOHMM states.

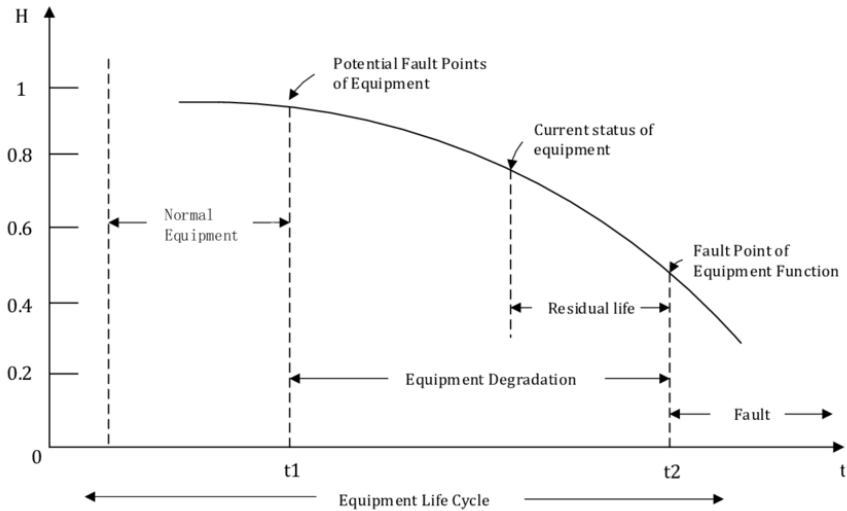

**Figure 3**: Health degradation curve of equipment, taken from Li et al. (2019)

### 3.5.1. Remaining Useful Life (RUL) estimation for Predictive Maintenance

Another additional benefit of using the IOHMM is RUL prediction per every cycle. The prediction followed the use of the Viterbi algorithm to predict the most optimal sequence of the state till that cycle and then based on the last observed state predict the next most probable state using the transition probabilities matrix. Using the Emission probabilities one can sample the sensor observations based on the predicted next state and append it to the previously seen observations sequence. This process is continued until the sequence predicted the next state to be the failure state as decoded previously. The total number of transitions to the failure state gives the Remaining Useful Life (RUL) at that particular cycle for each cycle the trend can be predicted. The algorithm is defined in Algorithm 3 elaborating on the implementation of the RUL estimation with the use of HMM state predictions.

## 4. Case Study: Predictive Maintenance of Turbofan Engines

Basic rotating components of a turbofan engine are arranged as shown in Figure 4. NASA Commercial Modular Aero-Propulsion System Simulation (C-MAPSS), turbofan engine degradation dataset Saxena and Goebel (2008) is

**Algorithm 3** RUL Estimation

**Input:**
$O$: observation sequence till cycle $\}t'$
$A$: Transition probability matrix
$B$: Emission probabilities
**for** *every cycle* **do**
   useful cycles = 0
   **while** *predicted state is not in failure state* **do**
     **for** 100 *iterations* **do**
       Predict state sequence using the Viterbi algorithm
       Select the most probable next state
       Sample sensor observations of the predicted state
       Append predicted state-to-state sequence
       Append sampled observation to sequence
       Add 1 to 'useful cycles'
     **end for**
     RUL = Average of the useful cycles
   **end while**
**end for**
RUL for each point = list of RULs
**Output:** RUL prediction

widely used in the community of predictive maintenance. The dataset consists of several engine units with multivariate time-series sensor readings and operating conditions discretized based on the flight cycles. Each unit observes some initial degradation at the start of the equipment failure, after which the health of the equipment degrades exponentially until it reaches a final failure state, hence, having a run-to-failure simulation. However, these degrees of wear are unknown. Recently, NASA published an updated version of the dataset Chao (2021) that records the real-time flight data and appends the operational history to the degradation modeling. This dataset additionally provides the ground truth values for the health state of the engine based on the component failure modes. The summary of the dataset used in this paper is shown in Table 1.

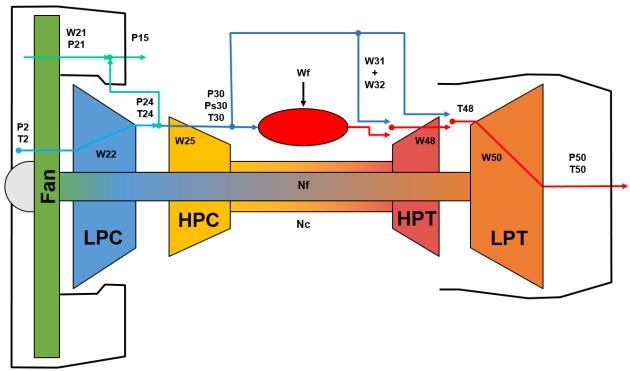

**Figure 4:** Turbofan engine and sensor configuration used by NASA C-MAPSS simulation Chao (2021)

The proposed derived methodology in Section 3.3 will be used for the use case of predictive maintenance of turbofan engines. In this section, the decision-making problem associated with optimal predictive maintenance is framed as an RL problem.

**Table 1**
Summary of the dataset used in this paper.

| NAME | OPERATING CONDITIONS | FAILURE MODES | GROUND TRUTH |
|---|---|---|---|
| FD001 | 1 | 1 | No |
| FD002 | 6 | 1 | No |
| FD003 | 1 | 2 | No |
| DS01 | NOT SPECIFIED | 1 | Yes |

## 4.1. Environment Dynamics and Modeling

The DRL framework for predictive maintenance proposed in Ong, Niyato and Yuen (2020) considers three actions as a general methodology for any decision-making maintenance model; *hold* [2], *repair*, and *replace*. The constraints can be the maintenance budget, and the objective function can be the maximum uptime of the equipment. We propose a general framework for modeling such environments with state transitions based on the actions selected under stochastic events (uncertainty of failure, and randomness of replacement by new equipment) at any state, as illustrated in Figure 5. "*Hold*" transitions the current state to the next state in time, under uncertainty of ending up in a failure state. "*Repair*" transitions the current state of the equipment back in its life cycle to an arbitrary state as defined by the type of repair or through some standards either from experience, reference manual, or history of data. "*Replace*" transitions the current state of the current equipment to the initial state of the next equipment (introducing randomness), however, if the equipment reaches its final (failure) state regardless of the action chosen; the equipment must be replaced now.

Although the general framework presented in Figure 5 includes three actions (hold, replace, and repair), for simplicity and due to the lack of data for repair actions in most cases, the action space presented in this paper, consists of just two actions (hold or replace). Algorithm A.1 of Appendix A defines the modeling of such stochastic dynamics in terms of RL used within the Open AI gym environment Brockman et al. (2016).

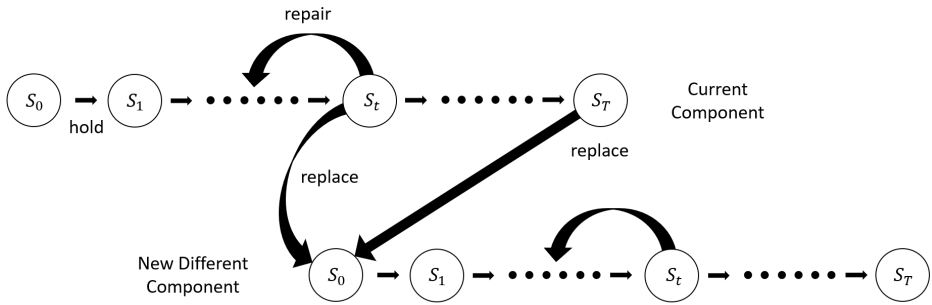

**Figure 5:** Dynamics of the model of the environment

Further, the dynamics of such a system have been illustrated in the context of Dynamic Programming (DP) and Reinforcement Learning (RL). Section 4.1 shows the graph structure of decision-making (hold, repair, replace) under uncertainty with probability ($\mathbb{P}$) for a possible maintenance application. It demonstrates the state-action pairs with each large circle representing a state at time t and small black circles representing the action with uncertain transition states. With the action of hold, there is a probability of either ending up in a failure state ($S_T$) at time $T$ or transitioning to the next operating cycle ($S_{t+1}$). The superscript in the figure specifies the component/machine's unique identifier. If the action of replacement is selected or if the replacement is performed after the component or machine has failed (corrective maintenance) then the new component will be selected with a different identifier starting from its initial life cycle ($S_0$). However, if the action of repair is chosen then the component or the machine does not change and it transitions to an arbitrary previous state in time of its life cycle depending on the repair type and the cycle continues from that state.

---
[2]The action of "hold" means that the agent neither suggests replacing nor repair and the system is healthy enough for the next operating cycle.

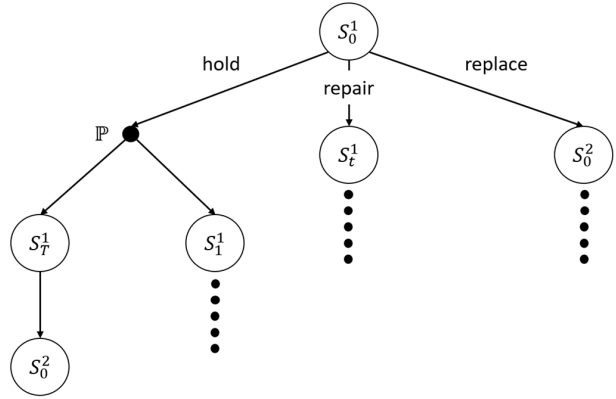

Figure 6: Environment Dynamics

## 4.2. Reward Formulation

For the maintenance decision having only replacement or hold actions, a dynamic reward structure has been formulated as shown in Equation (9) from Skordilis and Moghaddass (2020a). In this equation $c_r$ is the replacement cost, $c_f$ is the failure cost, $t$ is the current cycle, $T_j$ is the final (failure) cycle, and $r_t$ is the immediate reward. This cost formulation maintains the trade-off between early replacement and replacement after failure.

$$r_t = \begin{cases} 0, & a_t = \text{Hold} \quad \& \; t < T_j, \\ -\frac{c_r}{t}, & a_t = \text{Replace} \quad \& \; t < T_j, \\ -\frac{c_r + c_f}{T_j}, & a_t = \text{Hold} \quad \& \; t = T_j, \\ -\frac{c_r + c_f}{T_j}, & a_t = \text{Replace} \quad \& \; t = T_j. \end{cases} \quad (9)$$

## 4.3. Evaluation Criteria

To evaluate the performance of the RL agent, these two numerical values were chosen:

1. Cost
2. Average remaining useful life

### 4.3.1. Cost

The average optimal total return ($\widetilde{Q^*}$) serves as a numeric value used and compared with the upper and lower bounds of cost for such conditions Skordilis and Moghaddass (2020a).

*Ideal Maintenance Cost (IMC)* serves as the lower bound and the ideal cost in such maintenance applications. It is the incurred cost when the replacement action is performed one cycle before the failure, as shown in Equation (10). In this equation $N$ denotes the number of equipment used for evaluation, $\mathbb{E}(T)$ is the expected failure state of the equipment.

$$\phi_{IMC} \approx \frac{N \cdot c_r}{N \cdot (\mathbb{E}(T) - 1)} \approx \frac{N \cdot c_r}{\sum_{j=1}^{N} (T_j - 1)} \quad (10)$$

*Corrective Maintenance Cost (CMC)* serves as the upper bound and the maximum cost in such maintenance applications. It is the incurred cost when the replacement action is performed after the equipment has failed as shown in Equation (11).

$$\phi_{CMC} \approx \frac{(c_r + c_f)}{\mathbb{E}(T)} \approx \frac{N \cdot (c_r + c_f)}{\sum_{j=1}^{N} T_j} \quad (11)$$

*Average Optimal Cost* ($\widetilde{Q^*}$) is the average cost that the agent receives as its performance on the test set as shown in Equation (12). In this equation $r(s, a)$ denotes the immediate reward as formulated in Equation (9), $Q^*(s', a')$ denotes the optimal action value of the next state-action pair, and $\gamma$ is the discount factor.

$$\widetilde{Q^*}(s, a) = \frac{1}{N} \sum \left[ r(s, a) + \gamma \max_{a'} Q^* \left( s', a' \right) \right] \quad (12)$$

### 4.3.2. Average Remaining Useful Life ($\widetilde{RUL}$) before replacement

It quantifies; how many useful cycles are remaining on average when the agent proposes the replacement action. Ideally, it should be "*1*" according to the defined criteria.

## 5. Experimental Setup

The baseline systems defined in this paper are distinguished and designed by varying each of these four stages: (i) input, (ii) feature engineering, (iii) RL architecture, and (iv) output.

*System 1: Baseline*

(i) Raw sensor data as the input, (ii) Standard normalization as the feature engineering module, (iii) DNN as the RL architecture, and (iv) Action policy as the output; as shown in Figure 7.

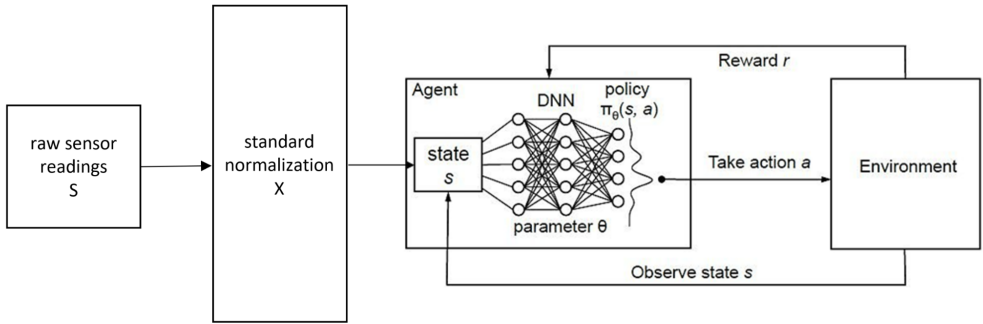

**Figure 7**: HMM posterior probabilities as the input to DRL.

*System 2: Baseline + Operating Conditions*

(i) Raw sensor data and operating conditions as the input, (ii) Standard normalization as the feature engineering module, (iii) DNN as the RL architecture, and (iv) Action policy as the output; as shown in Figure 8. It is used to set the failure cost to be used for the rest of the experiments.

*System 3: Baseline + HMM*

(i) Raw sensor data as the input, (ii) MinMax normalization and HMM as the feature engineering module, (iii) DNN as the RL architecture, and (iv) Action policy, and event-based unsupervised clustering and interpretation as the output; as shown in Figure 9.

*System 4: Baseline + Operating Conditions + IOHMM*

(i) Raw sensor data and operating conditions as the input, (ii) MinMax normalization and IOHMM as the feature engineering module, (iii) RNN as the RL architecture, and (iv) Action policy, RUL estimation, and unsupervised clustering and interpretation based on events at output; as shown in Figure 10. Its significance is to determine the optimal number of IOHMM states to be used in the experiments. Implementation of IOHMM is done through a library Yin and Silva (2017). This baseline uses the output of the IOHMM (probability distribution) as the input to the DRL agent, whereas SRLA uses the raw data as the input to the DRL agent during the state of specialization.

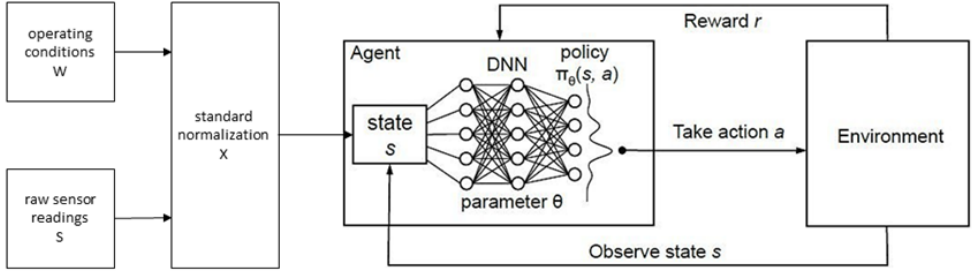

Figure 8: HMM posterior probabilities as the input to DRL.

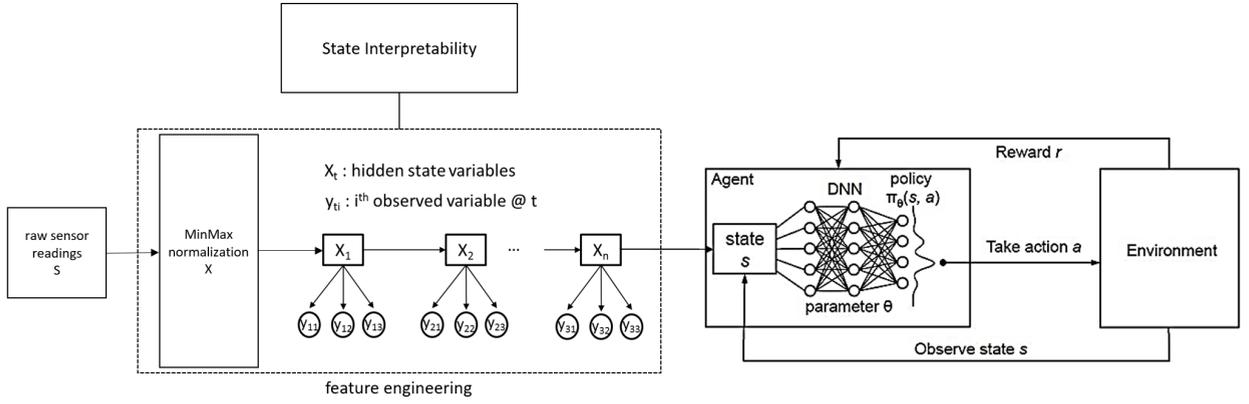

Figure 9: HMM posterior probabilities as the input to DRL.

## 5.1. Training Parameters
The summary of the DL framework within the RL architectures is as follows: (a) Deep Neural Network (DNN) consisting of a total of 37,000 training parameters and fully-connected (dense) layers with 2 hidden layers having 128 and 256 neurons, respectively, with ReLU activation. (b) Recurrent Neural Network (RNN) consists of a total of 468,000 training parameters and fully connected (LSTM) layers with 2 hidden layers having 128 and 256 neurons, respectively. The output layer consists of the number of actions the agent can decide for decision-making with linear activation. The parameters of the DRL agent are as follows: discount rate = 0.95, learning rate = 1e-4, and the epsilon decay rate = 0.99 is selected with the initial epsilon = 0.5.

## 5.2. Convergence Criteria
### 5.2.1. IOHMM
For IOHMM, Expectation Maximization (EM) tolerance and 1000 training epochs were used as the convergence criterion, whichever occurred earlier. EM tolerance was set to 1e-5. It corresponds to the difference between the last value of EM and the current value when it becomes less than the set tolerance, the training stops.

### 5.2.2. DRL
For DRL, the training loss was used as the convergence criterion and it was set to 1e-4 or 10,000 training episodes, whichever occurred earlier. The training loss threshold corresponds to the mean squared error between the predicted state-action value function ($Q$) and the actual $Q$ as estimated by the target $Q$ function.

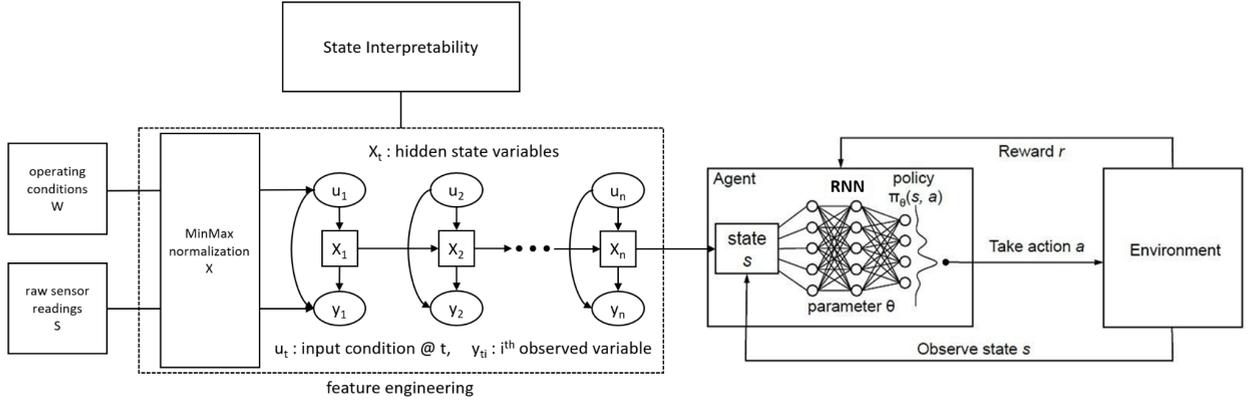

**Figure 10:** IOHMM posterior probabilities as the input to DRL.

### 5.3. Setting the Hyperparameters for the Models

This section describes the experiments used to determine the hyperparameters (i) cost of failure ($c_f$) and (ii) number of HMM/IOHMM states. The effectiveness of the architectures has been evaluated as described in Section 4.3. The data set used for this part of the experiment is FD001, which is split into an 80:20 (train:test) ratio.

#### 5.3.1. Calculating the cost of failure

The reward function (Equation (9)) for the RL agent requires the specification of *cost of failure* ($c_f$) and *cost of replacement* ($c_r$). However, the NASA C-MAPSS data set does not specify these parameters. To fix these values, we train Baseline 1 using a range of different $c_f$, while fixing $c_r$ and then comparing and identifying the $c_f$ that minimizes the average of total optimal cost per episode ($\widetilde{Q}^*$). We used System 2 because this system has the baseline architecture while at the same time using the full set of input parameters available in the dataset. $c_r$ is fixed (100) and the comparison is based on the different $c_f$ values of 25, 500 and 1000, as shown in Table 2. It was observed that as $c_f$ increases, $\widetilde{Q}^*$ becomes closer to the ideal cost, and, at the same time, the number of failed units decreases to 0%. However, the agent becomes more cautious, suggesting replacement action earlier in the lifetime of the engine; thereby, increasing the average remaining cycles. In the context of predictive maintenance of safety-critical systems, it is more important to avoid failure at the expense of replacing equipment a few cycles before its remaining useful life. Therefore, $c_f$ of 1000 was chosen for the rest of the experiments. Table 2 also shows the results of the optimal action policy learned by the agent through System 1. As a comparison, it can be concluded that the additional information of the operating conditions helps the model to learn a better maintenance policy.

#### 5.3.2. Calculating the number of hidden states

System 3 was used to find the number of states of the HMM/IOHMM model that maximizes the likelihood of our state space and the performance of the DRL through an iterative process. We evaluated the performance of the model as the number of states varied between 10, 15, and 20 states. The model trained through HMM and IOHMM gives the posterior probability distribution for every state as shown in Section 3.3 and Section 3.3, which is then fed as an input to the DRL agent to be able to learn the optimal maintenance (replacement) policy. The experiment was carried out on the test set using the failure cost of 1000 and with the same parameters as the previous experiment for a better evaluation. 15 states of the HMM/IOHMM showed better performance results than the rest, and so in the rest of our experiment, we use 15 as the number of states for the HMM and IOHMM model. Furthermore, the model with the HMM outperforms System 1 and System 2 as shown in Table 2.

## 6. Experiment 1: Comparison of IOHMM-DRL with Baseline and Prior Work

Until now, the dataset used just consisted of 1 operating condition, however, in real-world cases, there exists multiple operating conditions where HMM would fail. Therefore, to adapt to a more general architecture, an Input-Output Hidden

**Table 2**
Comparative evaluation and hyperparameter search.

| | Fail Cost | Avg Q* | IMC | CMC | Average Remaining Cycles | Failed Units |
|---|---|---|---|---|---|---|
| | | | System 2 | | | |
| | 25 | 0.54 | 0.45 | 0.56 | 2.4 | 45% |
| | 500 | 0.61 | 0.45 | 2.68 | 7.5 | 5% |
| | 1000 | 0.49 | 0.45 | 4.92 | 7.0 | 0% |
| | | | System 1 | | | |
| | 1000 | 0.51 | 0.45 | 4.92 | 12.0 | 0% |
| | | | System 3 | | | |
| HMM/IOHMM States | | | | | | |
| 5 | | 0.60 | 0.45 | 4.92 | 44.8 | 0% |
| 10 | | 0.54 | 0.45 | 4.92 | 24.2 | 0% |
| 15 | | 0.49 | 0.45 | 4.92 | 6.8 | 0% |
| 20 | | 0.53 | 0.45 | 4.92 | 20.2 | 0% |
| 30 | | 0.55 | 0.45 | 4.92 | 28.5 | 0% |

**Table 3**
Comparison of the proposed methodology with baseline systems and Skordilis and Moghaddass (2020a) on dataset FD002.

| Methodology | $\widetilde{Q}^*$ | IMC | CMC | IMC/$\widetilde{Q}^*$ | Failure | Average Remaining cycles | Interpretations |
|---|---|---|---|---|---|---|---|
| System 1 | 2.10 | 0.64 | 7.02 | 0.30 | 20% | 5.9 | No |
| System 2 | 6.87 | 0.64 | 7.02 | 0.09 | 90% | 2.6 | No |
| System 3 | 7.02 | 0.64 | 7.02 | 0.09 | 100% | 0.0 | Yes |
| System 4 | 0.77 | 0.64 | 7.02 | 0.83 | 0% | 23.0 | Yes |
| PF + DRL [15] | 2.02 | 1.93 | 20.80 | 0.96 | 0% | - | No |
| SRLA | 0.69 | 0.64 | 7.02 | 0.94 | 0% | 6.4 | Yes |

Markov Model (IOHMM) is used instead of the HMM. Data set FD002 with 6 operating conditions is used in this experiment for the comparative evaluation with baselines and prior work Skordilis and Moghaddass (2020a).

### 6.1. Comparative Evaluation and Results

As will be seen in Section 7.2, the IOHMM can align its states and state transitions with the relevant health states of the engine; however, the definition and alignment of the states are not fine enough to replace the engine with just one cycle before the failure. Therefore, DRL is used to refine the granularity after state distribution based on IOHMM, resulting in a hierarchical model. To evaluate the performance, the results are compared with the four baseline systems and the Particle Filtering (PF) based-DRL (*Bayesian particle filtering*) framework proposed by Chen et al. (2003). In their experiments Chen et al. (2003) used 80 engines as the training set and 20 as the test set out of 260 engines. However, the engines were selected randomly; therefore, an exact comparison with the average agent cost could not be made. Therefore, the ratio of the Ideal Maintenance Cost (IMC) to the average agent cost ($\widetilde{Q}^*$) was compared in Table 3. As shown, System 4 (IOHMM) performs better than System 1 (sensor readings), 2 (sensor readings + operating condition), and 3 (HMM). IOHMM-DRL framework, on the other hand, outperforms all the baseline systems and has a comparative performance with the PF + DRL methodology with the added benefits of interpretability.

**Table 4**
Feature (sensor) Importance.

| STATE: 9 | STATE: 14 |
|---|---|
| feature 5: -12.497 | feature 5: 4.211 |
| feature 6: -3.873 | feature 6: 0.268 |
| feature 7: -5.984 | feature 7: 0.175 |
| feature 8: 0.463 | feature 8: 19.697 |
| feature 9: -7.529 | feature 9: 0.325 |
| feature 10: -12.737 | feature 10: 3.973 |
| feature 11: -3.454 | feature 11: 0.153 |
| feature 12: -5.651 | feature 12: 0.097 |
| feature 13: 4.036 | feature 13: -3.555 |

**Table 5**
Feature to sensor description.

| FEATURE | SENSOR | DESCRIPTION |
|---|---|---|
| feature 5 | $P_{30}$ | PRESSURE AT HPC OUTLET |
| feature 8 | $epr$ | ENGINE PRESSURE RATIO |
| feature 10 | $phi$ | FUEL FLOW : PRESSURE (HPC) |
| feature 13 | $BPR$ | BYPASS RATIO |

## 7. Experiment 2: Interpretations Based on the Hidden States

Data sets FD001, FD003, and DS01 are used in this section using the IOHMM for event-based hypothesis and state interpretations. The experiments performed here are to address the question of whether the introduction of the hidden states can help towards interpretability.

### 7.1. Interpretability - Failure Event Hypothesis

Due to the unavailability of the ground truth for other state mappings in FD003, just the failure states (last cycle state) were mapped in this experiment. Each failure state in the dataset is annotated with one of the 2 failure modes (HPC and fan degradation); however, the ground truth for the engines corresponding to which failure mode is not provided. Figure 11(a) plots state distributions for each data point based on the hidden states of the IOHMM on FD003. The data points are collected from the sensor readings of every engine per cycle; reduced to 2D features through Principal Component Analysis (PCA) for visualization. It was hypothesized that each of these state clusters defines a particular event. Analyzing the failure states revealed two IOHMM states that corresponded to the failure event (state 9 and 14) as visualized in Figure B.1 of Appendix B, which might be based on the two failure modes. To validate this hypothesis, the analysis was repeated with FD001 as shown in Figure 11(b), where there is only one failure mode defined in the description of the data set, and this analysis showed that only one IOHMM state was observed to be the failure state for each engine as shown in Figure B.2 of Appendix B. This suggests that it is possible to map IOHMM states to failure events within the health state of the equipment.

Using the feature importance methodology described in Section 3.5, features (sensor readings) with a relatively higher score (based on feature importance) were selected from each class (failure states depicted by IOHMM). Further, the corresponding actual sensor information and description were extracted from Saxena and Goebel (2008) as described in Table 5. From the background information from the sensor descriptions, it was observed that the sensor importance for the two different IOHMM states showed a concrete failure event interpretation that corresponded to the failure described in the data set (HPC and Fan degradation), as hypothesized in Table 6.

### 7.2. Interpretability - State Decoding and Mapping

The second version of the NASA C-MAPSS data set Chao (2021) was used here to evaluate the state interpretability of IOHMM throughout the engine life, a subset of which is shown in Figure 12, where the red trend represents the

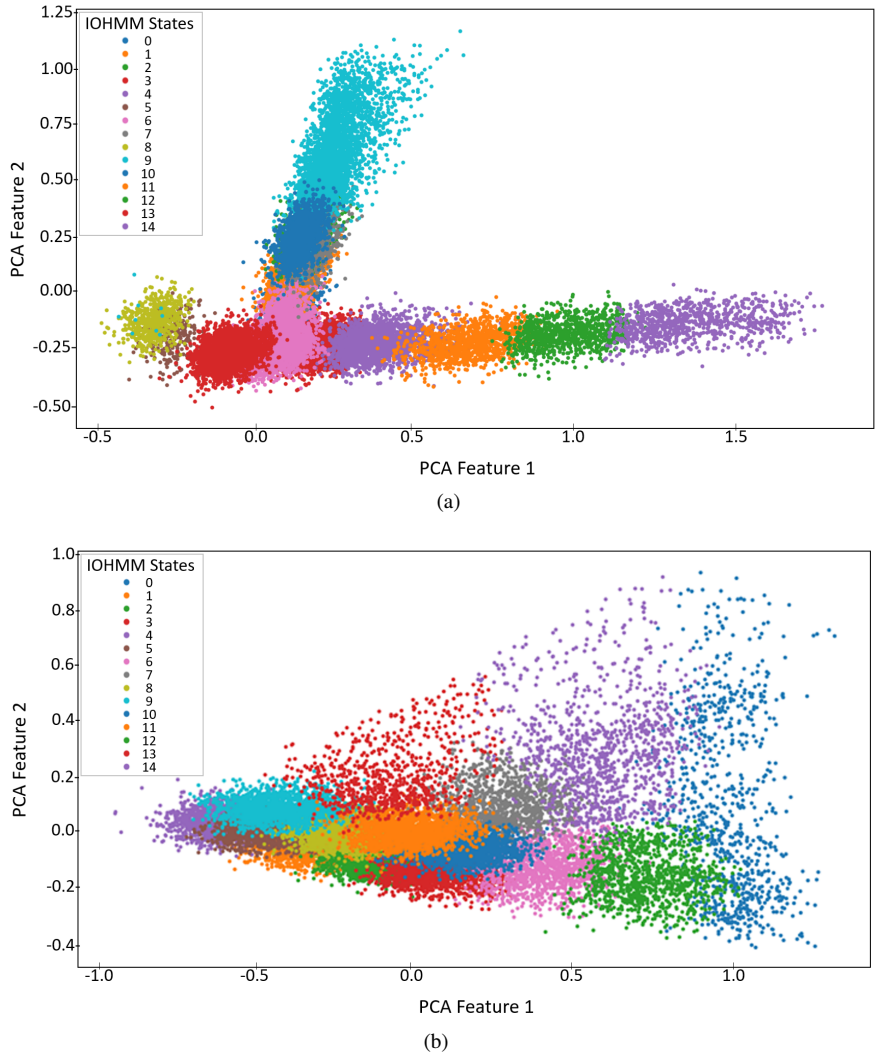

**Figure 11:** IOHMM states clustering for (a) FD003 and (b) FD001.

**Table 6**
Sensor importance to failure event hypothesis.

| HMM STATE | IMPORTANT SENSOR READING | FAILURE EVENT HYPOTHESIS (INTERPRETATION) |
|---|---|---|
| 9 | $BPR$ | FAN DEGRADATION |
| 14 | $P_{30}$, $epr$, $phi$ | HPC DEGRADATION |

IOHMM state prediction based on Equation (8). The data set has the ground truth values of the engine's state per cycle, the Boolean health state value is represented by the blue line with state 1 being healthy and 0 being unhealthy, the RUL is represented by the yellow line, and the green curve represents the health degradation curve. Based on the reference health degradation curve from Figure 3, and the range of IOHMM states observed during those conditions, we were able to associate different IOHMM states with different equipement conditions as shown in Table 7.

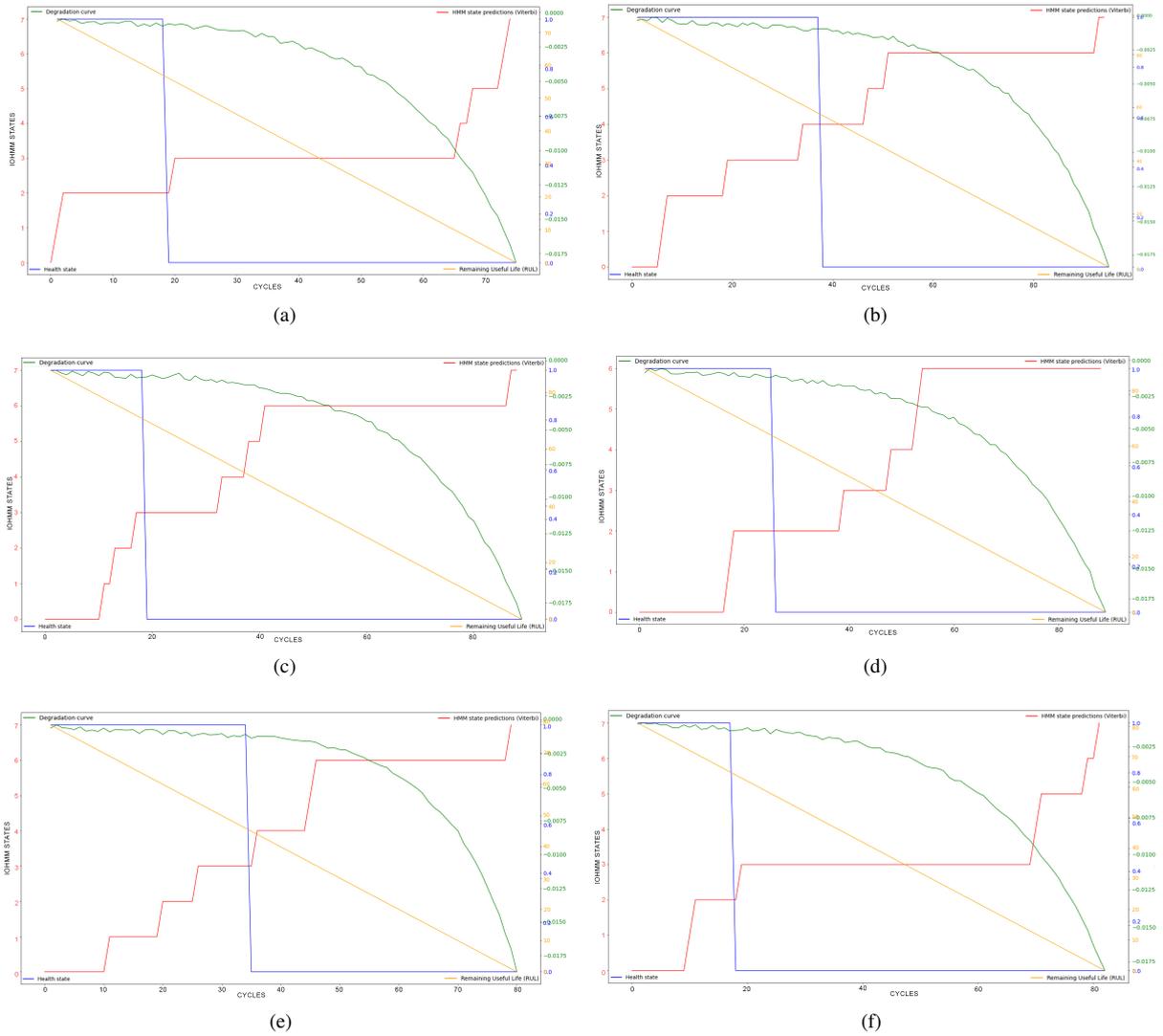

**Figure 12:** State decoding and mapping for dataset DS001.

## 7.3. Remaining Useful Life (RUL) Estimation

From Section 3.5.1, it was defined that the total number of transitions to the failure state gives the Remaining Useful Life (RUL) at that particular cycle. For each cycle the trend can be predicted as shown in Figure 13 following the algorithm as defined in Algorithm 3 from Section 3.5.1.

## Conclusion

In this paper, we proposed a novel approach for the use of deep reinforcement learning in safety-critical systems, called the Behavioral Cloning-Based Specialized Reinforcement Learning Agent (BC-SRLA). BC-SRLA addresses several challenges associated with the use of RL in these types of industries, including the need for continuous interaction with the environment, the preference for interpretable white-box systems, the uncertainty of machine learning techniques, and the inefficiency of deep reinforcement learning in complex and high-dimensional state-action spaces. The proposed hierarchical architecture combines the advantages of probabilistic modeling and model-free reinforcement learning with the added benefits of interpretability. It is activated in specific situations, such as abnormal conditions or when the system is near-to-failure and is initialized with a baseline policy using policy cloning. The effectiveness of the BC-SRLA was demonstrated through a case study in maintenance applied to turbofan engines, where it was compared to the state-of-the-art and other baselines and found to be superior.

**Table 7**
HMM state interpretability to equipment conditions.

| Equipment condition | HMM states |
|---|---|
| Normal equipment | 0 - 2 |
| Potential fault point of equipment | 2 - 4 |
| Failure progression | 4 - 6 |
| Fault point of equipment function | 6 - 7 |
| Failure | 7 |

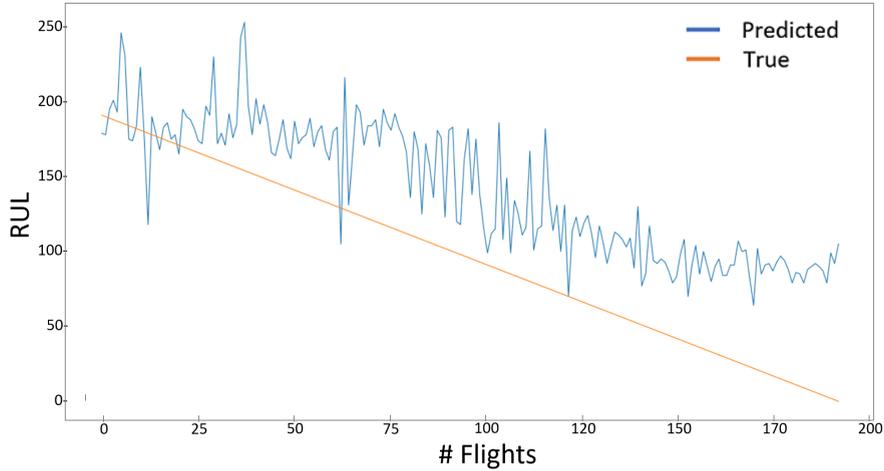

**Figure 13:** Remaining Useful Life estimation

## Future Work

There are several areas for future work in the development and implementation of the BC-SRLA. One potential direction is to further test and validate it in other safety-critical industrial applications and environments. Additionally, the interpretability of the model could be further improved through the development of more human-understandable explanations of the decision-making process. Another area for future work is the optimization of its performance, including the development of more efficient training methods and the incorporation of techniques to address the data inefficiency of deep reinforcement learning. Finally, further research could be conducted on the integration of the BC-SRLA with conventional methodologies in safety-critical industries and comparing it with the current state-of-the-art methodologies.

## A. *Algorithms*

---

**Algorithm A.1** Environment Modeling
---
  **Input:**
  $S_t = s_0, ..., s_t, ..., s_T$: state space
  $A_t = a_0, ..., a_n$: action space
  $R_t(s_t, a_t)$: reward given current state and chosen action
  **repeat**
    step in environment and sample observed state and reward
    **if** *hold* **then**
      **if** equipment has reached the failure state **then**
        reward of failure
        end of the episode
        replace to new different equipment and observe $S_0^{m+1}$
      **else**
        reward of hold
        increasing the age of the equipment by one step
        observed next state: $S_{t+1}$
      **end if**
    **else if** *replace* **then**
      **if** equipment has reached the failure state **then**
        reward of failure
      **else**
        reward of replacement
      **end if**
      end of the episode
      replace to new different equipment and observe $S_0^{m+1}$
    **end if**
    **Output:** $S_{t+1}$ : next state, $R_t$: reward, end of episode
  **until** all states have been observed

---

**Algorithm A.2** Approximate Reinforcement Learning

**Input:**
$T_{max}$: epochs
$\epsilon_0$: initial exploration parameter
$\gamma$: discount factor
$\alpha$: learning rate
$S_t = s_0, ..., s_t, ..., s_T$: state space
$A_t = a_0, ..., a_n$: action space
$R_t(s_t, a_t)$: reward given current state and chosen action
**for** $t = 1$ **to** $T_{max}$ **do**
  **if** *training* **then**
    $\epsilon = \epsilon_0$
  **else**
    $\epsilon = 0$
  **end if**
  **if** *exploration* $< \epsilon$ **then**
    choose action randomly
  **else**
    Choose action with maximum q-value
  **end if**
  find $S_{t+1}$ and $R_t$ given the chosen action
  approximate $Q$ for actions in current states $\hat{Q}(s_t, a_t)$
  sum approximated $Q$ for chosen actions $\mathbb{E}\{\hat{Q}(s_t, a_t)\}$
  approximate $Q$ for actions in next state $\hat{Q}(s_{t+1}|s_t, a_t)$
  compute $Q^*(s_{t+1}|s_t, a_t) = max(\hat{Q}(s_{t+1}|s_t, a_t))$
  $Q_{target} = R_t + \gamma(Q^*(s_{t+1}|s_t, a_t))$
  update $\hat{Q}$ by MSE between target and previous value;
  $\hat{Q}(s_t, a_t) \rightarrow \hat{Q}(s_t, a_t) + \alpha(\frac{1}{n}\Sigma(Q_{target} - \hat{Q}(s_t, a_t)))^2$
  total reward = $\Sigma(R_t)$
  $S_t = S_{t+1}$
  decay epsilon by a defined percentage
  **if** last state **or** end of episode **then**
    $Q_{target} = R_t(s_t, a_t)$
    break the loop
  **end if**
**end for**
**Output:** $\hat{Q}^*(s_t, a_t)$

## B. *Figures*

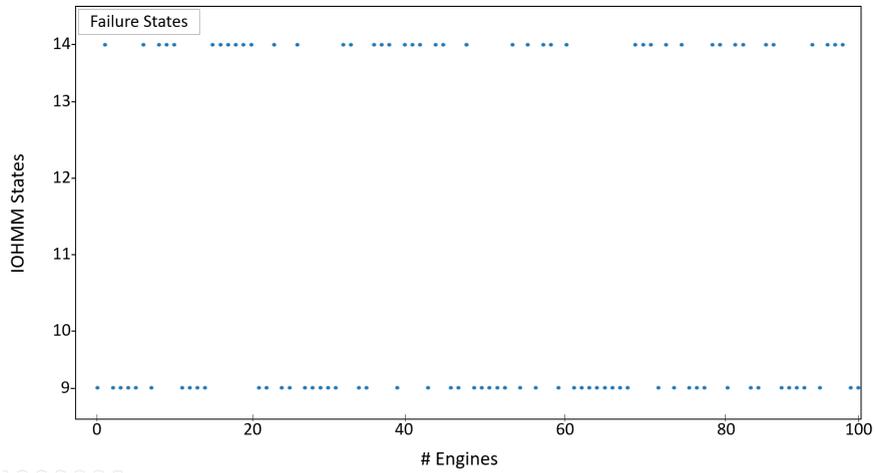

**Figure B.1:** States decoding and mapping for dataset FD003.

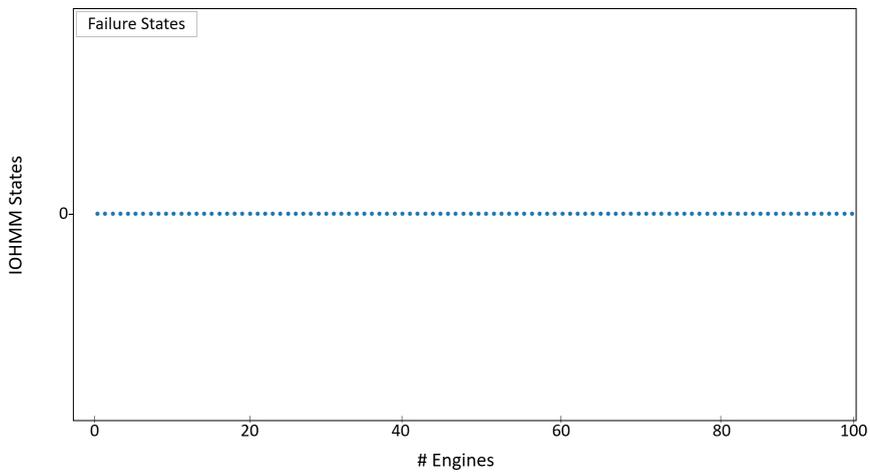

**Figure B.2:** States decoding and mapping for dataset FD001.